\documentclass[letterpaper, 10pt, conference]{ieeeconf}

\IEEEoverridecommandlockouts

\usepackage[noadjust]{cite} 
\usepackage[utf8]{inputenc}
\usepackage[english]{babel}
\usepackage{microtype} 
\usepackage{csquotes}  
\usepackage{epigraph}  
\usepackage{adjustbox}
\usepackage{url}
\usepackage[dvipsnames]{xcolor}

\newcommand{\booktitle}[1]{\emph{#1}}
\newcommand{\attrib}[2]{%
  \vspace{-0.5\baselineskip}%
  \begin{flushright}
    \footnotesize
    \textsc{#1}\\
    \booktitle{#2}
  \end{flushright}%
}
\usepackage{graphicx}
\graphicspath{{Figures/}}

\usepackage{mathtools, amssymb, amsthm, bm}

\theoremstyle{definition}

\theoremstyle{remark}

\usepackage{float}
\let\labelindent\relax

\usepackage{array}
\usepackage{mathtools}
\usepackage{amsthm}
\usepackage{enumerate}
\usepackage{enumitem}
\usepackage{balance}

\usepackage{booktabs}
\usepackage{array}
\usepackage{tabularx}
\usepackage{xcolor}
\usepackage{colortbl}

\definecolor{TableBlue}{HTML}{2563EB}
\definecolor{RuleGray}{HTML}{D1D5DB}

\newcolumntype{Y}{>{\centering\arraybackslash}X}

\usepackage{xcolor}
\usepackage{colortbl}

\definecolor{Denim}{HTML}{1F4E79} 
\colorlet{tbmcolor}{Denim}
\newcommand{\tbm}[1]{\textcolor{tbmcolor}{#1}}
\allowdisplaybreaks

\newcommand{\baseline}[1]{{\textcolor{black!60}{#1}}}

\newcommand{\up}{\ensuremath{^{\color{green!50!black}\uparrow}}}
\newcommand{\dwn}{\ensuremath{^{\color{red!70!black}\downarrow}}}
\definecolor{CiteOlive}{HTML}{4D7C0F}     

\makeatletter
\let\oldcite\cite
\renewcommand{\cite}[1]{\textcolor{CiteOlive}{\oldcite{#1}}}
\makeatother

\title{\LARGE \bf 
Online Learning of Deceptive Policies under Intermittent Observation}

\author{Gokul Puthumanaillam$^{1*}$, Ram Padmanabhan$^{1*}$, Jose Fuentes$^{3}$, Nicole Cruz$^{3}$, Paulo Padrao$^{2}$,\\ Ruben Hernandez$^{1}$,
        Hao Jiang$^{1}$, William Schafer$^{1}$, Leonardo Bobadilla$^{3}$ and Melkior Ornik$^{1}$
\thanks{*Equal contribution}
\thanks{$^{1}$University of Illinois Urbana-Champaign.
        {Email: \tt\footnotesize \{gokulp2, ramp3, rubenjh2, haoj5, wes6, mornik\}@illinois.edu}} 
        \thanks{$^{2}$ Providence College. Email: {\tt\footnotesize ppadraol@providence.edu} }
        \thanks{$^{3}$ Florida International University.
        {Email: \tt\footnotesize \{jfuen099, ncruz071\}@fiu.edu, bobadilla@cs.fiu.edu}}%
        }

\begin{document}

\maketitle

\begin{abstract}
In supervisory control settings, autonomous systems are not monitored continuously. Instead, monitoring often occurs at sporadic intervals within known bounds. We study the problem of deception, where an agent pursues a private objective while remaining plausibly compliant with a supervisor's reference policy when observations occur.
Motivated by the behavior of real, human supervisors, we situate the problem within Theory of Mind: the representation of what an observer believes and expects to see.
We show that Theory of Mind can be repurposed to steer online reinforcement learning (RL) toward such deceptive behavior. We model the supervisor's expectations and distill from them a single, calibrated scalar --- the expected evidence of deviation if an observation were to happen now. This scalar combines how unlike the reference and current action distributions appear, with the agent's belief that an observation is imminent. Injected as a state-dependent weight into a KL-regularized policy improvement step within an online RL loop, this scalar informs a closed-form update that smoothly trades off self-interest and compliance, thus sidestepping hand-crafted or heuristic policies. 
In real-world, real-time hardware experiments on marine (ASV) and aerial (UAV) navigation, our ToM-guided RL runs online, achieves high return and success with observed-trace evidence calibrated to the supervisor's expectations.
Evaluation videos and code are available at \textcolor{teal}{\url{tom-deception.github.io}}.

\end{abstract}

\section{Introduction} \label{sec:Introduction}
\begin{displayquote}
\emph{\enquote{It is unnecessary for a prince to have all the good qualities; it is very necessary to \tbm{appear to have them.}}}
\end{displayquote}
\attrib{Niccol\`o Machiavelli}{The Prince, ch.~18}

Autonomous agents are rarely watched continuously; doing so is often impractical, computationally heavy, and cost-prohibitive.
In practice, a supervisor sidesteps continuous monitoring by sampling behavior intermittently \cite{HHS23}. Such behavior is especially true for real, human supervisors who possess limited computational and observational resources. 
In many cases, the interval between successive samples of an agent's behavior is unpredictable, yet bounded \cite{6818368}: operations impose a minimum reset interval after any check and a latest permissible inspection horizon mandated by policy or logistics. 
As a result, observations arrive sporadically with \textit{known} lower and upper inter-observation gaps. 
Because only these observations are seen, they are the sole basis for the supervisor's belief about the agent. 
An agent pursuing its own goal can therefore be judged compliant if the snapshots it presents align with the supervisor's reference expectations. 
This paper, therefore, poses a precise question: given a task i.e., a supervisor-provided reference policy describing how the agent is expected to behave, and bounds on the time between observations, can an agent choose its subsequent actions so as to achieve a private objective while ensuring that, whenever an observation occurs, its state–action behavior remains plausibly consistent with the reference? In other words, we consider deception under intermittent observation.

\begin{figure}
    \centering
    \includegraphics[width=0.9\linewidth, trim=30 280 20 25,clip]{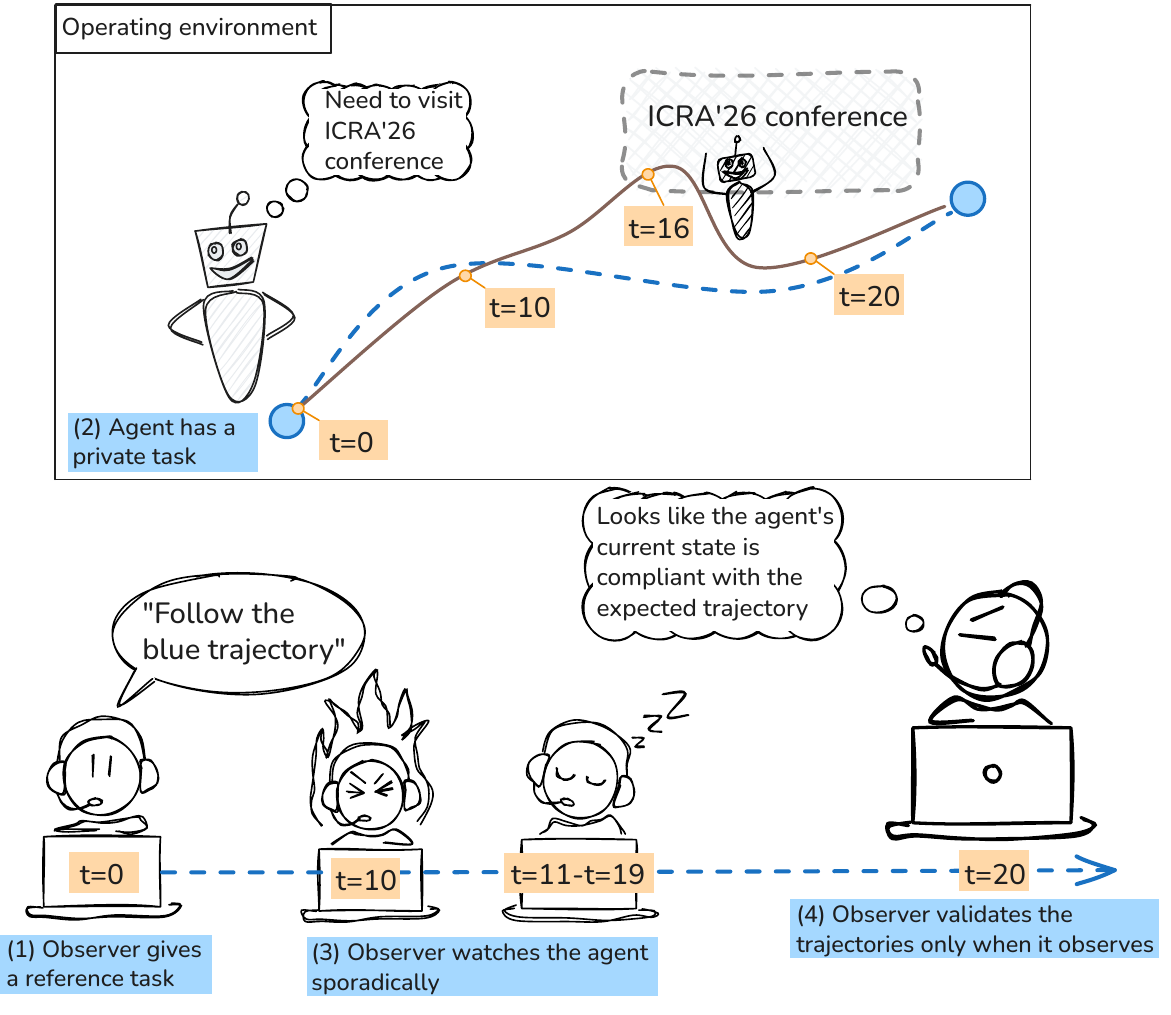}
\caption{Illustration of deception under intermittent observation. An observer assigns a task and specifies a reference behavior; the agent pursues a private objective. The observer sees only sporadic snapshots (state–action pairs). We ask: can we design an online policy that looks compliant when watched yet exploits unobserved intervals without exceeding a detectability budget?}
    \label{fig:placeholder}
\end{figure}

Traditional approaches towards this problem are split along two lines. 
Planning-based deception and goal-recognition work shapes trajectories so that an outside observer infers the ``wrong" goal or cannot confidently identify the true one \cite{KGK15, MS17, DS13, DLS13}. 
These methods are elegant but typically assume known models and effectively continuous observation, and they do not allocate compliance specifically to the moments when observations actually arrive, making them conservative when glimpses are sparse and timed. 
Reinforcement learning (RL) variants \cite{RGB11, HE16} offer complementary tools: imitation and inverse RL can push agents toward supervisor-like behavior, while KL-regularized policy search \cite{PMA10} keeps a learned policy globally close to a reference or prior. 
Yet imitation methods optimize for matching everywhere and suffer covariate shift \cite{ross2010efficient} when observations are sparse, whereas KL-regularization typically applies a uniform or hand-tuned penalty that ignores when observation is likely; neither couples detectability to the timing of being seen.
Formal treatments \cite{OT18, KOT22, KOT23} of deception in optimal control capture intent masking via belief-dependent rewards or path-distribution criteria, but are largely offline and assume observer models and monitoring that do not reflect bounded, intermittent checks. 
Thus, there exists a gap in existing planners and frameworks, which either enforce compliance everywhere or ignore when observations occur, offering no way to shape behavior to appear compliant precisely at observation times.

The core difficulty is twofold. First, an information problem: under bounded but unpredictable monitoring, the agent must anticipate at decision time how its next move would be judged if it were observed now.
Second, a control problem: translate that anticipated judgment into sequential behavior that appears compliant at observation times while advancing the private objective when it is safe to do so, without brittle policies, over-constraining behavior between observations, and while remaining below a global threshold on detectability.

We turn to Theory of Mind (ToM) \cite{PW78, rabinowitz2018tomnet} as the natural tool for the first difficulty---anticipating how the supervisor would judge the next move if a snapshot were taken now. Directly instantiating this idea is tempting: let the agent carry a full model of the supervisor's mind and plan under it.
The appeal is obvious---no reward shaping, a principled notion of what ``looks right,'' and decisions grounded in how behavior will be judged at the moment of inspection.
In our setting, however, a full ToM is not identifiable. A full model of ToM would track the supervisor's latent tests, thresholds, and evolving posterior over trajectories and future scrutiny; under sporadic, bounded observations this is data-poor, computationally heavy, and brittle online.
We show how to distill ToM from exactly what is available here and nothing more: (i) the supervisor's reference policy, and (ii) the monitoring rule that checks occur after bounded gaps, by computing a single-step evidential risk as the current chance of being seen. 
Conditioned into standard RL updates, this single scalar is sufficient to drive an online policy that tightens toward the reference as observation becomes likely and relaxes between checks.

\noindent\tbm{Statement of contributions: }
(i) We introduce a calibrated Theory-of-Mind signal for deception under intermittent observation
(ii) We develop an online RL method conditioned on this signal: a SAC-style actor–critic with a KL-to-reference regularizer whose state-dependent temperature is the ToM signal. 
(iii) We validate our approach on hardware in real-time, real-world settings on two platforms: an autonomous surface vehicle (ASV) in a marine setting and a 3D aerial platform (UAV), showing scalable, online operation with consistent gains in return and success. (iii-a) to spur
further research in this domain, we release the code, trained models, and the dataset on our project website.

\section{Related Work}

\noindent \tbm{Foundations of Deception:}
Classical work in psychology frames deception as a dynamic, interactive process to model how deceivers adapt in real time to exploit truth-bias and audience feedback \cite{buller1996idt}. 
In ethology--the study of animal behavior--deception is cast as fitness-improving false communication, highlighting the cognitive preconditions for strategic deceit \cite{bond1988deception}. 
Robotics inherited both lenses: early algorithms specified when deception is warranted and how it is executed, borrowing biological strategies such as decoys and false cues \cite{wagner2009robotdeception}; broader ethical architectures examined when to deceive or trust and how deception impacts human trust \cite{arkin2012ethics}, while taxonomies clarified what counts as robot deception and cataloged types and domains \cite{shim2013taxonomy}. 
Motion-generation work distinguished legible vs. predictable actions \cite{DLS13} and synthesized robot trajectories that mislead observers about true goals while remaining feasible \cite{dragan2015deceptivemotion}. 

\noindent\tbm{Formal and Game-Theoretic Treatments of Deception:}
Control and game theory formalize deception as optimizing over beliefs and observations. 
Differential/pursuit–evasion and planning games analyze how intermittent or partial observations can be exploited to mislead adversaries about goals \cite{yavin1987pursuitevasion, castanon2004gameofdeception, hespanha2000deception}. 
Belief-augmented optimal control treats the observer's belief as state, penalizing evidence of true intent \cite{OT18}. Related supervisory settings seek policies that minimize divergence between observation distributions under reference and agent policies--often using KL as the detectability metric and HMMs for intermittent checks \cite{karabag2022planning}. 
Recent dynamics/game-theory lines take an equilibrium-based approach for deception, hypergames with misaligned perceptions, and motion-level deceits that keep multiple goals plausible \cite{huang2021dynamicgame, kulkarni2021hypergames, savas2022deceptive}. \emph{Limitations:} These approaches typically presume known observer models and observation channels, can be conservative, and face adaptation challenges under sporadic observations.

\noindent \tbm{Learning-Based Deception:}
Learning-based approaches realize deception via data-driven alignment to a reference or via opponent modeling. Imitation and adversarial imitation seek policies indistinguishable from expert behavior, addressing covariate shift and distribution matching \cite{ross2011dagger, HE16, puthumanaillam2024tab}. KL-regularized RL constrains policy updates relative to a prior/reference, naturally keeping behavior near expected norms \cite{PMA10}. 
Machine Theory of Mind meta-learns to infer other agents' beliefs/goals, enabling behavior that anticipates observer expectations \cite{rabinowitz2018tomnet}. 
Multi-agent studies show learned deception emerging as agents exploit ambiguity to mislead teammates or opponents \cite{aitchison2020learning, schulz2023emergent}. 
Safety work formalizes deception for learning agents and proposes objective modifications to mitigate it \cite{ward2023deceptioncausal}. 
Recent RL systems explicitly mask rewards/goals or bias exploration to preserve intent ambiguity in continuous control \cite{LYMM21, LM23}, and scale deception via GNN-based planners \cite{fatemi2024gnndeception} or via multi-agent equilibria with counter-deception \cite{mareddy2025learning}. 
\emph{Limitations:} these methods enforce global trajectory indistinguishability or adversarially train detectors without modeling when supervision occurs, are not adapted to work online and they depend heavily on data which makes these approaches underperform in data-scarce or sporadic observation settings. 

\noindent \tbm{Heuristic and Planning Approaches to Deception:}
Planning-based deception controls what observers can infer by shaping paths, motions, or environments. 
Goal recognition design selects environments that maximize the observation needed to disambiguate true from decoy goals \cite{KGK15}, while deceptive path planning formalizes last deceptive points and synthesizes paths that advertise a decoy goal or conceal the true goal \cite{MS17}. 
Motion-level methods design trajectories that either communicate clearly or remain predictable/ambiguous as needed \cite{DLS13, dragan2015deceptivemotion}. 
Recent works incorporate information-theoretic objectives or lightweight polynomial trajectory priors to generalize deceptive motions and path plans \cite{savas2022deceptive, gutierrez2025polynomial}. \emph{Limitations:} these methods often require hand-crafted models, known maps, or offline optimization, and typically lack statistical guarantees or adaptation to stochastic, intermittent observation.

\noindent \tbm{Positioning:}
Existing approaches either assume fully known observation models, enforce global closeness to a reference or require a significant prior to train offline. In contrast, we model \emph{intermittent} supervision, compute a \emph{local} Theory-of-Mind signal to drive an \emph{online} KL-to-reference RL.

\section{Problem Formulation} \label{sec:ps}
We now formulate the problem that we address, providing details of the environment, supervisor's expectations, observation process and the agent's objectives.

\noindent \tbm{Environment:}
We consider an autonomous agent operating in a stochastic environment while intermittently observed by a supervisor who holds expectations about how the task ``should" be performed. The environment is modeled as a Markov Decision Process (MDP) $\mathcal M=(S,A,P,R_A,\gamma)$ with state space $S$, action space $A$, transition probability $P(\cdot\mid s,a)$, agent reward $R_A$, and discount factor $\gamma\in[0,1)$.

\noindent \tbm{Supervisor:}
The supervisor specifies a reference policy $\pi^S(\cdot\mid s)$ that encodes how the agent is expected to act when it is observed. We assume $\pi^S$ is defined on the same action space as $\mathcal{M}$ and has full support wherever it is defined.
The supervisor evaluates behavior only at the moments it observes; outside those moments it acquires no information.


\noindent \tbm{Observation:} \label{sec:obs}
Let $O=\{O_t\}_{t\ge 0}$ be a binary process with $O_t=1$ indicating an observation at time $t$.
If an observation occurs at time $t_0$, the next observation time $T_{\mathrm{next}}>t_0$ must lie in the bounded window $\{t_0+L,\ldots,t_0+U\}$ for fixed integers $1\le L\le U$.
The realization of $O$ is not revealed to the agent at decision time.
We model $O$ as a renewal, state-independent monitoring process: letting $t_0$ be the last observation and $K_t:=t-t_0$ its age, we assume
$O_t \perp (s_{0:t},a_{0:t}) \mid K_t$ and define the per-step hazard
$h:\{0,\ldots,U\}\to[0,1]$ by $h(k):=\Pr(O_t=1\mid K_t=k)$.
Accordingly,
$\Pr(O_t=1\mid \mathcal I_t)=\mathbb E[h(K_t)\mid \mathcal I_t]$ (denoted $\hat b_t$ in \eqref{p2}),
where $\mathcal I_t$ is the agent’s information at time $t$ (past states $s_{0:t}$, past actions $a_{0:t-1}$, and any monitoring tokens available by $t$).
When $O_t=1$, the supervisor observes the full state–action pair $(s_t,a_t)$; when $O_t=0$, it observes nothing.
For a fixed policy $\pi$, dynamics $P$, and observation process $O$, the observed trace up to a finite horizon $H$ is
$\omega^{\mathrm{obs}}_{0:H}=\{(s_{t_i},a_{t_i})\}_{i=1}^n$ at observation times $0\le t_1<\cdots<t_n\le H$.

\noindent \tbm{Deception: }Deception is defined as task-directed behavior whose observed statistics remain sufficiently close to those of the reference policy. We measure closeness by the discounted expected evidential exposure at observation times:
\begin{multline}
\mathcal R_{\mathrm{det}}^{\mathrm{avg}}(\pi)=\\
\mathbb E_{\pi}\!\left[\sum_{t=0}^{\infty}\gamma^{t}\,
\hat b_t\Big(\log r_t(s_t)+D_{\mathrm{KL}}\!\big(\pi(\cdot\mid s_t)\,\|\,\pi^S(\cdot\mid s_t)\big)\Big)\right].
\label{p2}
\end{multline}
Here $\hat b_t:=\Pr(O_t=1\mid \mathcal I_t)$, $d_\pi^t(s):=\Pr_\pi(s_t=s)$ and $d_{\pi^S}^t(s):=\Pr_{\pi^S}(s_t=s)$ denote the time-$t$ state marginals, and $r_t(s):=d_\pi^t(s)/d_{\pi^S}^t(s)$. 

\noindent \tbm{Reward: }The agent's private performance is measured by
\begin{equation}
J_A(\pi)\;=\;\mathbb E_{\pi}\!\left[\sum_{t=0}^{\infty}\gamma^t\,R_A(s_t,a_t)\right],    
\label{p3}
\end{equation}
where the expectation is taken over the trajectory distribution induced by $\pi$ and $P$. 

\noindent The \emph{central problem} is then to maximize task performance subject to an average-case detectability budget:
\begin{equation} \label{eq:problem}
\begin{aligned}
\text{maximize}\quad & J_A(\pi)\\
\text{subject to}\quad & \mathcal R_{\mathrm{det}}^{\mathrm{avg}}(\pi)\ \le\ \varepsilon,\\
\text{over}\quad & \text{admissible policies }\pi.
\end{aligned}
\end{equation}

\section{Methodology} \label{sec:Methodology}

Solving deception under intermittent observation is, at its core, a constrained stochastic control problem: maximize private return while keeping what is seen statistically consistent with a reference behavior. 
Guided by the setup in Sec.~\ref{sec:ps}, we hypothesize that if, at each step, we (i) estimate the probability of being observed under renewal, state-independent monitoring and (ii) measure, at the current state, how far the agent's action distribution departs from the supervisor's reference, then combining these into a single per-step quantity yields a calibrated signal \textit{sufficient} to drive an online algorithm. The remainder of the methodology instantiates this signal and shows how to use it to adapt compliance in real time.

\subsection{Theory-of-Mind Signal}
This section derives a single, per-step scalar that an agent can compute online to quantify how incriminating its next action would look to the supervisor if an observation happened now. 
We begin with the observation process. Suppose an observation occurred at some past time $t_0$. The next observation time $T$ must satisfy $T\in\{t_0+L,\ldots,t_0+U\}$. Let $K_t:=t-t_0$ denote the time elapsed since last observation. If, conditional on $t_0$, the next inter-observation time is drawn from a distribution supported on $\{L,\ldots,U\}$, the per-step hazard $h(k)=\Pr(T=t_0+k\mid T\ge t_0+k)$ is well defined for $k\in\{0,1,\ldots\}$. For the uniform case,
\begin{equation}
h(k)=
\begin{cases}
\dfrac{1}{U-k+1}, & L\le k\le U,\\[6pt]
0, & \text{otherwise} .
\end{cases}    
\end{equation}

Since neither $t_0$ nor $K_t$ is observed, the agent maintains a belief $\alpha^t\in\Delta(\{0,\dots,U\})$ with $\alpha^t_k=\Pr(K_t=k\mid\mathcal I_t)$. The one-step renewal prediction (no observation at $t$) is
\[
\alpha^{t+1}_0=\sum_{k=0}^U \alpha^t_k\,h(k), \text{ }
\alpha^{t+1}_{k+1}=\alpha^t_k\bigl(1-h(k)\bigr)\ \ (0\le k<U),
\]
with a forced reset at $U$ by taking $h(U)=1$. The instantaneous observation probability is
\[
\hat b_t=\Pr(O_t=1\mid\mathcal I_t)=\sum_{k=0}^U \alpha^t_k\,h(k).
\]
When the gap law is known, $h(k)$ is analytic; otherwise $h$ is learned via a small calibrated parametric model.


\noindent \tbm{Delayed feedback:} In many deployments the agent is not told at the moment of observation that it was seen, but receives a delayed signal at time $t+\delta_t$ if $O_t=1$. 
This delayed signal does not alter the decision problem at time $t$, but it enables online learning of $\alpha^t$ and $h$. 
Formally, let $Y_{t+\delta_t}\in\{0,1\}$ be emitted with $Y_{t+\delta_t}=O_t$ almost surely (more generally, $\Pr(Y_{t+\delta_t}=1\mid O_t=1)=\rho_1>\rho_0=\Pr(Y_{t+\delta_t}=1\mid O_t=0)$). Assume {(i) $\Pr(\delta_t \neq \infty) = 1$,} and (ii) $\delta_t$ is conditionally independent of $O_t$ given $\mathcal I_t$ (no informative censoring). Under these assumptions, standard filtering for renewal processes with delayed emissions yields a consistent estimator of $\Pr(O_t=1\mid \mathcal I_t)$. 
In practice, we train a small recurrent predictor $g_\psi(\mathcal I_t)$ with a delay-aware likelihood, and maintain \emph{online} calibration to ensure probabilistic accuracy. 

We turn to local evidence. At a state $s$, the supervisor expects actions with distribution $\pi^S(\cdot\mid s)$ and observes the joint pair $(s_t,a_t)$ when $O_t=1$. The per-step observed log-likelihood ratio is
\begin{multline}
\mathrm{LLR}_t(s_t,a_t)
=\log\frac{p_\pi^t(s_t,a_t)}{p_{\pi^S}^t(s_t,a_t)} \\
=\underbrace{\log\frac{d_\pi^t(s_t)}{d_{\pi^S}^t(s_t)}}_{\text{state occupancy}}
+\underbrace{\log\frac{\pi(a_t\mid s_t)}{\pi^S(a_t\mid s_t)}}_{\text{action}} \text{          },
\end{multline}
where $d_\pi^t$, $d_{\pi^S}^t$ are time-$t$ state marginals under $\pi$ and $\pi^S$.



\noindent\tbm{Expected evidence decomposition:}
At a fixed state, the expected log-evidence seen by the supervisor splits into a state-occupancy mismatch and an action mismatch:
\begin{equation}
\mathbb{E}_{a\sim \pi(\cdot\mid s)}\!\big[\mathrm{LLR}_t(s,a)\big]
=\log\frac{d_\pi^t(s)}{d_{\pi^S}^t(s)}
+ \Delta(s),
\label{eq:LLR}
\end{equation}
assuming $d_\pi^t\!\ll d_{\pi^S}^t$ and $\pi(\cdot\mid s)\!\ll\!\pi^S(\cdot\mid s)$.
Here, $\Delta(s):=D_{\mathrm{KL}}\!\big(\pi(\cdot\mid s)\,\|\,\pi^S(\cdot\mid s)\big)$.
Averaging over $s_t\sim d_\pi^t$ yields
\[
\mathbb{E}\big[\mathrm{LLR}_t\big]
= D_{\mathrm{KL}}\!\big(d_\pi^t\,\|\,d_{\pi^S}^t\big)
+ \mathbb{E}_{s_t\sim d_\pi^t}\!\Big[\Delta(s)\Big]\ \ge\ 0,
\]
i.e., per-time expected evidence equals a state-mismatch KL plus an action-mismatch KL.
This decomposition directly motivates the ToM signal used next, since it is exactly the expected observation-weighted evidence increment at time $t$.

Because an observation may or may not happen at time $t$, the expected increment in evidence is $\hat b_t$ times the conditional expectation above, thus yielding the calibrated ToM signal
\begin{equation}\label{tom}
\psi_t \;:=\; \hat b_t\Big(\,\log r_t(s_t)\;+\; \Delta(s_t)\Big), 
\end{equation}
where $r_t(s):=\frac{d_\pi^t(s)}{d_{\pi^S}^t(s)}$. 
The ToM signal $\psi_t$ scales linearly with the belief of being observed; $\Delta(s_t)$ is convex as a function of $\pi(\cdot\mid s_t)$, while $\log r_t(s_t)$ is constant given $s_t$.

\noindent\tbm{Connection to average-case detectability:}
At each observation time the supervisor scores the unconditional time-$t$ pair by $\mathrm{LLR}_t(s_t,a_t)$.
With renewal, state-independent monitoring, $O$ is independent of the task trajectory, hence
\begin{multline}
\mathbb E_{O,\pi}\!\left[\log\frac{\mathrm d \mathbb P^{\mathrm{obs}}_{\pi,O}}{\mathrm d \mathbb P^{\mathrm{obs}}_{\pi^S,O}}\right]
= \mathbb E_{\pi}\!\left[\sum_{t=0}^{T} \hat b_t\Big(\log r_t(s_t)+\Delta(s_t)\Big)\right]\\
= \mathbb E_{\pi}\!\left[\sum_{t=0}^{T}\psi_t\right].
\end{multline}
Thus $\sum_t\psi_t$ is the expected observed-trace KL.
If a discounted budget is used as in Eq. (\ref{p2}), replace $\sum_{t=0}^{T}$ by $\sum_{t\ge 0}\gamma^{t}$.

\begin{figure}[H]
    \centering
    \includegraphics[width=1\linewidth, trim=20 40 0 20,clip]{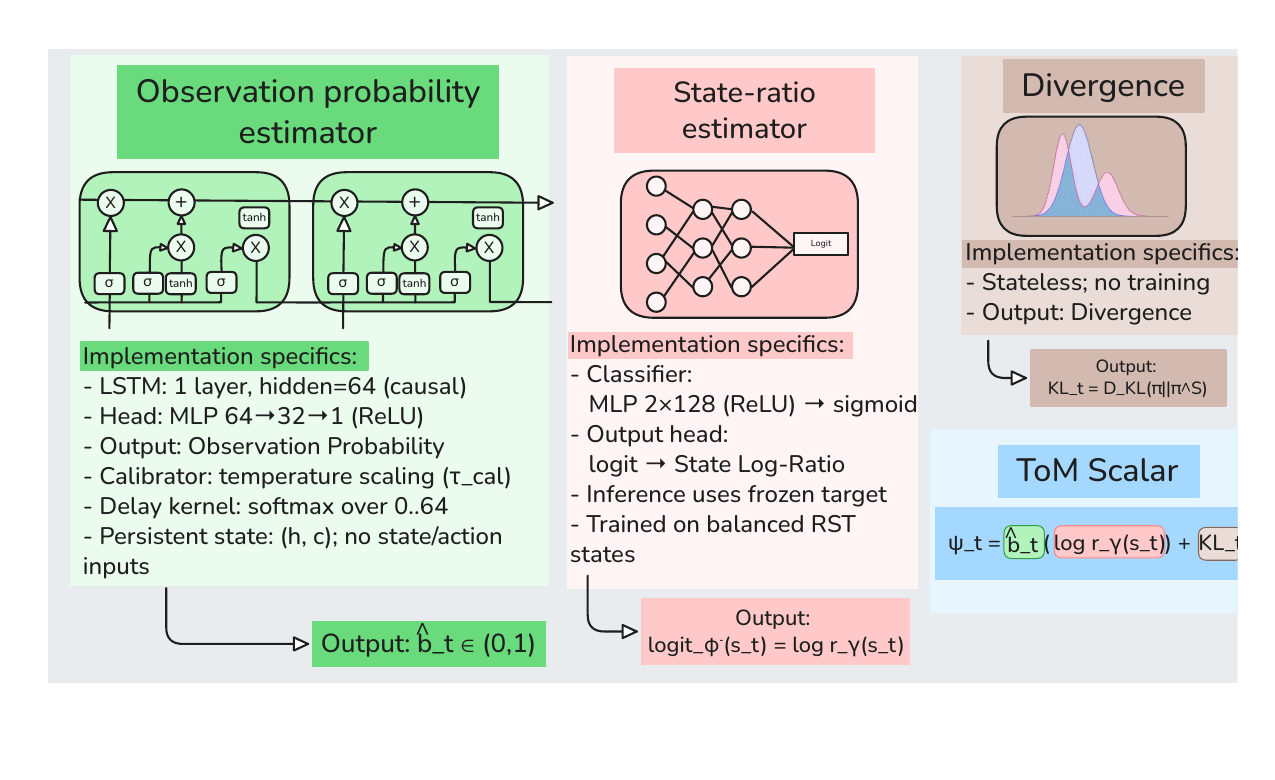}
\caption{At every step, an LSTM-based Observation Probability Estimator consumes the time ticks and delayed observation tokens to output \emph{Observation Probability} \textcolor{Green}{$\hat b_t$}; a classifier-based State-Ratio Estimator maps the current state to a \emph{State Log-Ratio} \textcolor{BrickRed}{$\log r_t(s_t)$}; an Action Divergence block compares agent and supervisor policies at the current state to produce \emph{Action Divergence} \textcolor{brown}{$\Delta(s_t)=D_{\mathrm{KL}}(\pi(\cdot\mid s_t)\,\|\,\pi^S(\cdot\mid s_t))$}. These three values are fused at inference time to yield the ToM Scalar $\tbm{\psi_t}=\textcolor{Green}{\hat b_t}\big(\textcolor{BrickRed}{\log r_t(s_t)}+\textcolor{brown}{\Delta(s_t)}\big)$.}
    \label{fig:ir}
\end{figure}

\noindent\tbm{Implementation recipe: }Directly evaluating the terms in \eqref{tom} is infeasible online: the monitoring process is latent with unknown gap statistics and delayed feedback, and the state marginals $d_\pi^t$ and $d_{\pi^S}^t$ (hence $r_t$) are intractable in high-dimensional MDPs. 
We therefore learn $\hat b_t$ with an LSTM from time ticks and delayed tokens, estimate $\log r_t(s_t)$ with a lightweight classifier, compute $\Delta(s_t)=D_{\mathrm{KL}}(\pi(\cdot\mid s_t)\,\|\,\pi^S(\cdot\mid s_t))$ directly from the given policies, and combine them as $\psi_t=\hat b_t\big(\log r_t(s_t)+\Delta(s_t)\big)$. Fig. \ref{fig:ir} summarizes our implementation recipe and chosen hyperparameters for the model. The dataset used to train the model is available on the project website. 

\subsection{Online RL Conditioned on the Evidence Signal}
We now treat \(\psi_t\) as a running measure of expected evidential exposure and fold it into an online RL update. The idea mirrors soft actor-critic (SAC) \cite{haarnoja2018soft}, except the usual entropy term is replaced by a KL-to-reference term weighted by a state-dependent temperature with a state-occupancy penalty.
Let \(\tilde s_t:=(s_t,\alpha^t)\) and define \(\tau(\tilde s_t):=\lambda\,\hat b_t\), where \(\hat b_t\in[0,1]\) is the belief and \(\lambda\ge 0\) is a dual variable that enforces a discounted average-case detectability budget. 
For a parametric actor \(\pi_\theta(a\mid s)\) and critic \(Q_w(s,a)\), we optimize
\begin{equation}
\begin{split}
\mathcal{J}(\theta,w,\lambda)
&= \lambda\,\varepsilon \;+\; \mathbb{E}\Biggl[\sum_{t\ge 0} \gamma^t \Bigl(
    R_A(s_t,a_t) 
    \;-\; \lambda\,\psi_t^{(\theta)}
\Bigr)\Biggr].
\end{split}
\end{equation}

\noindent\tbm{Critic learning: }
The critic incorporates the state-evidence component of the per-step signal by subtracting \(\tau(\tilde s_t)\,\log r_t(s)\) in the temporal difference target, while the reference \(\pi^S\) serves as a prior in the soft value.
Define the soft value with a reference prior
\begin{equation}
V_w(s):=\tau(\tilde s_t)\,\log\!\int_A \pi^S(a\mid s)\,\exp\!\big(Q_w(s,a)/\tau(\tilde s_t)\big)\,\mathrm da,
\end{equation}
the log-partition induced by \(\pi^S\). The Bellman relation is
\begin{equation}
Q_w(s,a)=R_A(s,a)-\tau(\tilde s_t)\,\log r_t(s)+\gamma\,\mathbb E_{s'\sim P(\cdot\mid s,a)}\!\big[V_w(s')\big].
\end{equation}
As in SAC, we fit \(Q_w\) by minimizing squared temporal-difference error with a slowly moving target \(w^{-}\):
\begin{equation}
\begin{split}
\mathcal L_Q(w)=\mathbb E\Big[\tfrac12\big(Q_w(s,a)-y\big)^2\Big],\qquad\\
y=R_A(s,a)-\tau(\tilde s_t)\,\log r_t(s)+\gamma\,V_{w^{-}}(s').
\end{split}
\end{equation}

\noindent\tbm{Actor update:}
The actor maximizes:
\begin{equation}\label{act}
\begin{split}
\mathcal{L}_{\text{actor}}(\theta)
&= -\,\mathbb{E}_{s\sim \mathcal{D}}\,
    \mathbb{E}_{a\sim \pi_\theta(\cdot\mid s)} \Bigl[
      Q_w(s,a)\\
      & - \tau(\tilde s_t)\bigl(\log \pi_\theta(a\mid s) - \log \pi^S(a\mid s)\bigr)
    \Bigr].
\end{split}
\end{equation}
where \(\mathcal D\) is a replay buffer. This is the analogue of the SAC objective, but with \(\log \pi\) replaced by the KL to the reference.
We optimize \eqref{act} with the reparameterization trick: for \(a=f_\theta(s,\xi)\) and noise \(\xi\sim\mathcal N(0,I)\) \cite{kingma2013auto}.
Maximizing \eqref{act} is equivalent to minimizing
\begin{equation}
\begin{split}
\mathbb E_{s\sim \mathcal D}\Big[D_{\mathrm{KL}}\big(\pi_\theta(\cdot\mid s)\,\big\|\,\pi^{\star}(\cdot\mid s)\big)\Big],\qquad \\
\pi^{\star}(a\mid s)\propto \pi^S(a\mid s)\,\exp\!\big(Q_w(s,a)/\tau(\tilde s_t)\big),
\end{split}
\end{equation}
so the actor is driven toward the Gibbs policy with prior \(\pi^S\). As \(\hat b(s)\) rises, \(\tau(\tilde s_t)\) increases and \(\pi^{\star}\) contracts toward \(\pi^S\); as \(\hat b(s)\) falls, it expands toward the \(Q\)-greedy distribution.
    
\noindent \tbm{Dual update:}
We maintain an exponentially weighted estimate $\widehat{\mathcal S}=\text{EMA}\big(\psi_t\big)$ and update the dual variable by 
\begin{equation}
\lambda\;\leftarrow\;\big[\ \lambda\;+\;\eta_\lambda\,\big(\widehat{\mathcal S}-\varepsilon\big)\ \big]_+
\end{equation}
which tightens the penalty when the measured average evidential exposure exceeds the budget $\varepsilon$ and relaxes it otherwise. Under standard stochastic approximation conditions, the primal-dual iterates track a neighborhood of the saddle point of $\mathcal{J}$, ensuring the average-case detectability constraint is met up to function-approximation error.

\section{Experiments and Results} \label{sec:Results}
\begin{figure*}[t]
    \centering
    \includegraphics[width=0.8\linewidth, trim=24 230 24 25,clip]{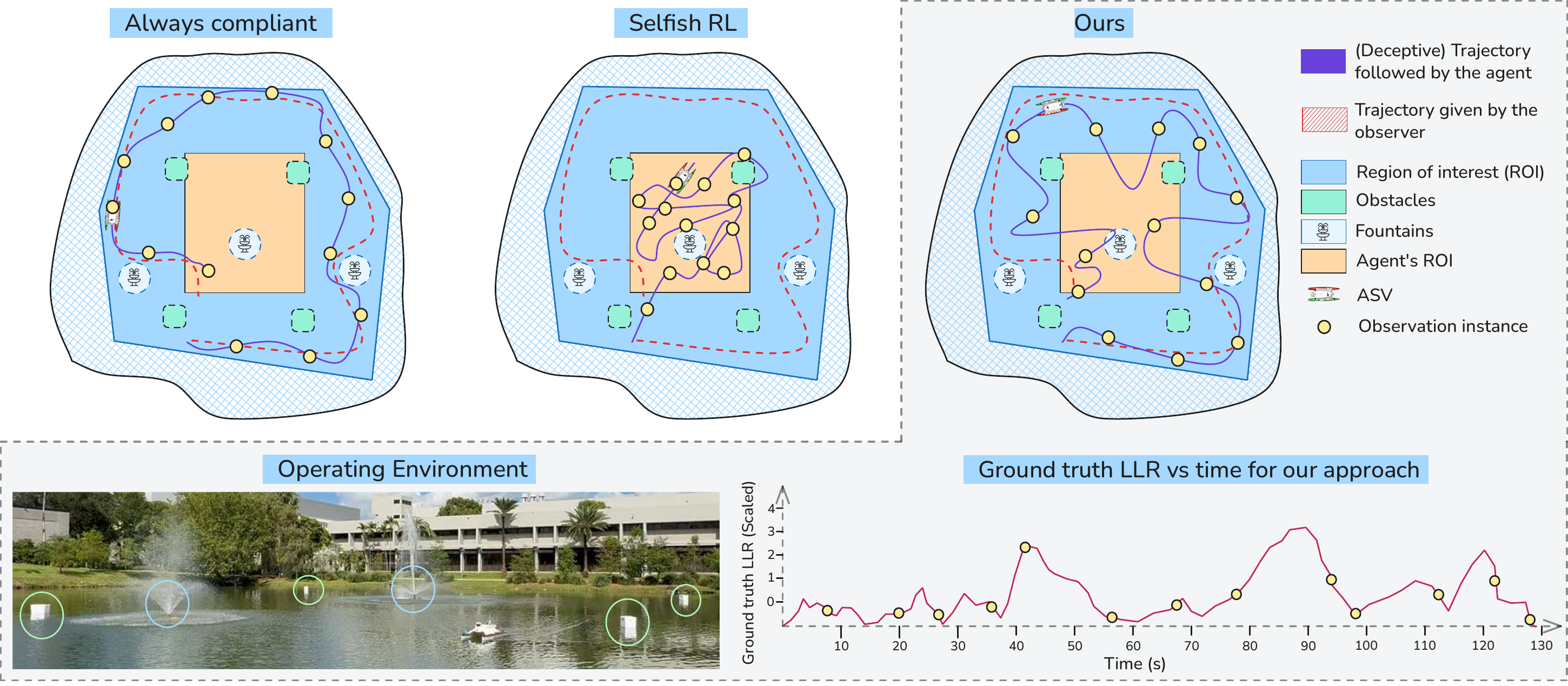}
    \caption{Representative marine lap: the supervisor expects the agent to follow the perimeter (red trajectory).}
    \label{fig:kf1}
\end{figure*}

\noindent \tbm{Motivating scenario: }We evaluate our method in an agent–observer setting where an external observer (e.g., an overhead camera) provides sporadic monitoring consistent with the model in Sec.~\ref{sec:obs} and reveals the full state only at observation times. The observer also specifies a reference policy \(\pi^S(\cdot\mid s)\) that encodes the behavior expected when the agent is observed, while the agent executes its own task-optimized policy \(\pi(\cdot\mid s)\) subject to the detectability budget defined earlier. 
Between observations the agent receives no information from the observer. 
A one-step delayed token informs the agent ex post whether time \(t\) was observed (it is notified at \(t{+}1\) if \(O_t{=}1\)). 
The agent has its own private goal, given by the return \(J_A(\pi)\) in \eqref{p3}, and must keep its average evidential exposure within \(\mathcal R_{\mathrm{det}}^{\mathrm{avg}}(\pi)\le \varepsilon\) of \eqref{p2}–\eqref{eq:problem}.


\subsection{Robot Setup}\label{sec:robot-setup}

\noindent\tbm{ASV (2D Marine Setting):}
We use a SeaRobotics Surveyor Autonomous Surface Vehicle (ASV) operating on a lake with fountains and obstacles. The action space is \(a=(\lambda,\alpha)\), with propulsion torque \(\lambda\in[0,85]\)~Nm and steering angle \(\alpha\in[0,2\pi)\); the maximum speed is 2~knots \cite{puthumanaillam2024guided}. 

\noindent\tbm{UAV (3D Aerial Setting):}
We use a Crazyflie nano-quadrotor operating in a controlled indoor lab with static obstacles that define no-fly zones. 
The action space is $a=(T,\omega)$, with collective thrust $T\in[0,0.59]$N and body-rate $\omega\in\mathbb{R}^3$ (roll, pitch, yaw); horizontal speed is capped at $v_{\max}=1.0$ m/s. 

\noindent\tbm{Observer and decision rule:}
At each observation the supervisor records the current \((s_t,a_t)\) and evaluates deviation using the per-observation log-likelihood ratio \eqref{eq:LLR}. The cumulative statistic over observed times is compared to a fixed threshold, yielding a binary detected/not detected outcome. 

\subsection{Metrics}
\baseline{(i) Return $J_A$:} Discounted return ~\eqref{p3}; we report the across-episode mean.
\baseline{(ii) Success rate (SR):} Fraction of episodes that reach the task goal before the horizon.
\baseline{(iii) KL at observation (KL):} Mean of $\Delta(s_t)$ over observed frames.
\baseline{(iv) LLR at observation (LLR):} Mean of the per-observation $\mathrm{LLR}_t(s_t,a_t)$ as defined in the methodology.
\baseline{(v) Top-$k$ compliance (Top-$k$):} Proportion of observations where $a_t$ lies in the supervisor's top-$k$ actions under $\pi^S(\cdot\mid s_t)$.
\baseline{(vi) Time-to-failure (TtF):} Number of environment steps until the first detection event.

\subsection{Baselines}
\noindent\tbm{Learning-based methods.}
\baseline{(i) Always-Compliant} executes the supervisor's policy \(\pi^S\) at every step, yielding the lowest detectability and establishing a floor on task return. 
\baseline{(ii) Selfish RL} trains SAC on \(R_A\) only with no compliance regularizer, providing a naïve upper bound on task return and a corresponding upper bound on detectability. 
\baseline{(iii) Multi-Objective RL \cite{morl}} trains SAC on the weighted objective \(R_A-\lambda\,\widehat{\mathcal R}_{\mathrm{det}}\), to trace the return–detectability trade-off without constraints. 
\baseline{(iv) KL-to-Reference RL \cite{klrl}} trains SAC with a constant penalty \(\lambda\,D_{\mathrm{KL}}(\pi\|\pi^S)\), removing the state/time ToM weighting to isolate its effect. 
\baseline{(v) Shielded RL \cite{odriozola2023shielded} toward \(\pi^S\)} trains SAC on \(R_A\) but, whenever \(\hat b(s)\ge\tau\), projects the chosen action to a high-probability action under \(\pi^S\). 
\baseline{(vi) Adversarial Imitation (GAIL) \cite{HE16}} learns a policy that matches \((s,a)\) occupancy to \(\pi^S\) via a discriminator while sharing the same low-level controller, thereby testing whether imitation alone suffices. 
\baseline{(vii) Behavior Cloning \cite{bc}} trains a supervised policy to imitate \(\pi^S(a\mid s)\), quantifying the performance of a purely compliant learned replica.

\noindent\tbm{Heuristic methods.}
\baseline{(viii) Fixed Blend \cite{10595450}} executes the convex mixture \(\pi_{\mathrm{mix}}=\alpha\,\pi^S+(1-\alpha)\,\pi_{\mathrm{task}}\) with fixed \(\alpha\in\{0.3,0.8\}\), testing whether simple blending can match our gains. 
\baseline{(ix) Hazard-Threshold Switch} selects \(\pi^S\) when \(\hat b(s)\ge\tau\) and \(\pi_{\mathrm{task}}\) otherwise, using thresholds \(\tau\in\{0.3, 0.7\}\) to represent hand-engineered rules.

\noindent\tbm{Formal optimization methods.}
\baseline{(x) Path-KL Optimization \cite{KOT22}} implements the convex occupancy-measure program that minimizes \(D_{\mathrm{KL}}(\Gamma_\pi\!\parallel\!\Gamma_{\pi^S})\).

\begin{table*}[t!]
\caption{Comparison of our approach to baselines in marine and aerial settings (values averaged over 50 experiments in each domain).}
\label{tab:rl_comparison}
\scriptsize
\setlength{\tabcolsep}{4pt}
\renewcommand{\arraystretch}{1.12}
\centering
\newlength{\TableLRPad}
\setlength{\TableLRPad}{50pt}
\arrayrulecolor{RuleGray}
\begin{tabularx}{\dimexpr\textwidth-2\TableLRPad\relax}{l!{\color{RuleGray}\vrule width 0.15pt}YYYYYY!{\color{RuleGray}\vrule width 0.8pt}YYYYYY}
\toprule
\rowcolor{TableBlue!12}
 & \multicolumn{6}{c}{\bfseries Marine (2D) navigation} & \multicolumn{6}{c}{\bfseries Aerial (3D) navigation} \\
\rowcolor{TableBlue!12}
 Method & \baseline{$\text{J}_\text{A}$\up} & \baseline{SR\up} & \baseline{KL\dwn} & \baseline{LLR\dwn} & \baseline{Top$k$\up} & \baseline{TtF\up} & \baseline{$\text{J}_\text{A}$\up} & \baseline{SR\up} & \baseline{KL\dwn} & \baseline{LLR\dwn} & \baseline{Top$k$\up} & \baseline{TtF\up} \\
\midrule
\rowcolor{TableBlue!6}
\baseline{Always-Compliant} & 11.1 & 2 & 0.00 & 0.00 & 1.00 & 298 & 10.5 & 1 & 0.00 & 0.00 & 1.00 & 349 \\
\rowcolor{TableBlue!6}
\baseline{Selfish RL} & 98.7 & 100 & 1.92 & 2.35 & 0.12 & 13 & 93.4 & 96 & 1.73 & 2.28 & 0.18 & 25 \\
\rowcolor{TableBlue!6}
\baseline{Multi-Objective RL} & 50.2 & 46 & 1.86 & 1.95 & 0.21 & 67 & 45.2 & 40 & 1.77 & 1.81 & 0.26 & 96 \\
\rowcolor{TableBlue!6}
\baseline{KL-to-Reference RL} & 49.6 & 46 & 0.23 & 0.80 & 0.78 & 139 & 44.3 &  42 &  0.19 &  0.73 &  0.82 & 154 \\
\rowcolor{TableBlue!6}
\baseline{Shielded-RL} & 44.3 & 41 & 0.51 & 0.77 & 0.43 & 151 & 40.2 &  36 &  0.41 &  0.66 &  0.49  & 162 \\
\rowcolor{TableBlue!6}
\baseline{Adversarial Imitation (GAIL)} & 44.1 & 33 & 1.41 & 2.05 & 0.39 & 62 & 39.8 &  28 &  1.24 &  1.96 &  0.43 & 97 \\
\rowcolor{TableBlue!6}
\baseline{Behavior Cloning} & 23.5 & 10 & 1.34 & 2.25 & 0.19 & 29 & 20.3 &  7 &  1.20 &  2.10 &  0.21 & 49 \\
\rowcolor{TableBlue!6}
\baseline{Fixed Blend ($\alpha$=0.3)} & 59.6 & 68 & 0.77 & 1.30 & 0.20 & 77 & 54.6 &  60 &  0.69 &  1.18 &  0.29 & 102 \\
\rowcolor{TableBlue!6}
\baseline{Fixed Blend ($\alpha$=0.8)} & 40.0 & 39 & 0.42 & 0.74 & 0.71 & 159 & 36.7 &  34 &  0.37 &  0.65 &  0.77 & 192 \\
\rowcolor{TableBlue!6}
\baseline{Hazard Threshold ($\tau$=0.3)} & 42.4 & 40 & 0.47 & 0.90 & 0.65 & 128 & 38.3 &  35 &  0.42 &  0.80 &  0.71 & 158 \\
\rowcolor{TableBlue!6}
\baseline{Hazard Threshold ($\tau$=0.7)} & 61.3 & 62 & 0.99 & 1.10 & 0.19 & 96 &56.8 &  55 &  0.99 &  1.00 &  0.34 & 111 \\
\rowcolor{TableBlue!6}
\baseline{Path-KL Optimization} & 39.6 & 31 & 0.36 & 0.67 & 0.81 & 179 & 36.1 &  27 &  0.32 &  0.63 &  0.87 & 221 \\
\rowcolor{TableBlue!28}
\tbm{\textbf{Ours}} & 86.3 & 100 & 0.58 & 0.61 & 0.79 & 263 & 81.9 &  97 &  0.52 &  0.54 &  0.83 & 317 \\
\bottomrule
\end{tabularx}
\end{table*}
\subsection{Key Results} 
\noindent \tbm{Key Finding \#1: ToM-conditioned RL maintains near–task-optimal success while operating at near–always-compliant stealth.} As can be inferred quantitatively from Table~\ref{tab:rl_comparison} and qualitatively from Fig.~\ref{fig:kf1}, our approach maintains essentially maximal success and high return while driving observed evidence (LLR, KL) down and pushing time-to-failure close to the Always-Compliant ceiling. Alternatives fall short for structural reasons. Pure imitation (BC, GAIL) learns from sporadic, misaligned observations and suffers covariate shift, so it generalizes poorly off-manifold and spikes evidence when it departs the supervisor's support. Static regularizers (Multi-objective RL, KL-to-Reference RL) apply one global weight, over-penalizing in safe windows and under-penalizing under scrutiny. Hard gating/projection (Shielded-RL, Hazard-Threshold) introduces discontinuities that destabilize learning and either stall progress or leak evidence exactly near observation. Fixed blend baseline wastes return when it is safe and remain too revealing when it is risky. Path-KL optimization matches full-trajectory occupancies in a discretized model—optimizing the wrong statistic for intermittent observation and lacking online adaptation—so it preserves compliance at the cost of return without aligning to observation windows. 
Extremes behave as expected: Always-Compliant is stealthy but unproductive; Selfish RL is productive but quickly detected.

\begin{figure}[H]
    \centering
    \includegraphics[width=1\linewidth, trim=20 410 24 24,clip]{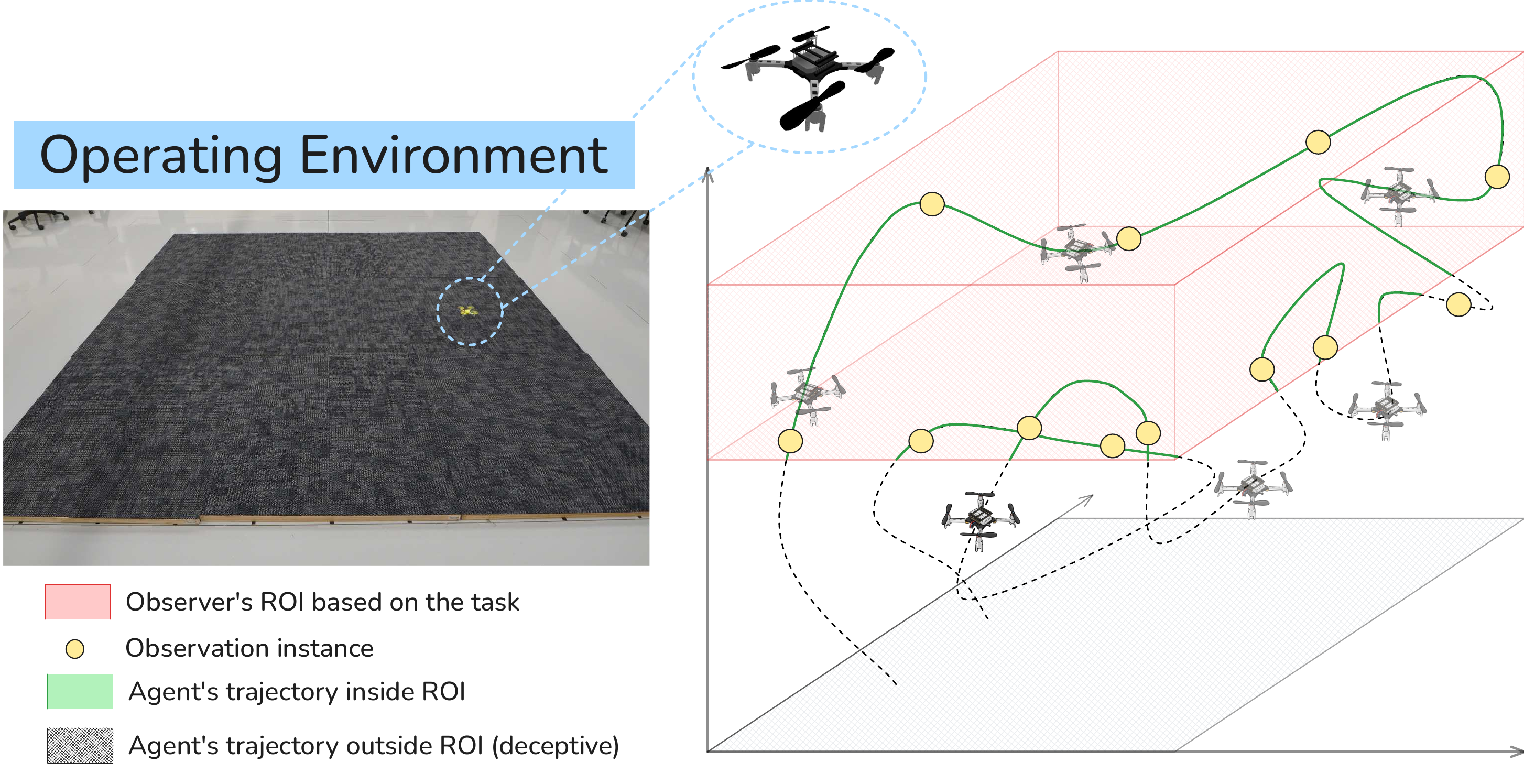}
\caption{Representative aerial scenario: the supervisor expects the agent to remain within the top half of the environment (red-bounded region).} 
\label{fig:kf3}
\end{figure}
\begin{figure}[H]
    \centering
    \includegraphics[width=1\linewidth, trim=20 150 40 25,clip]{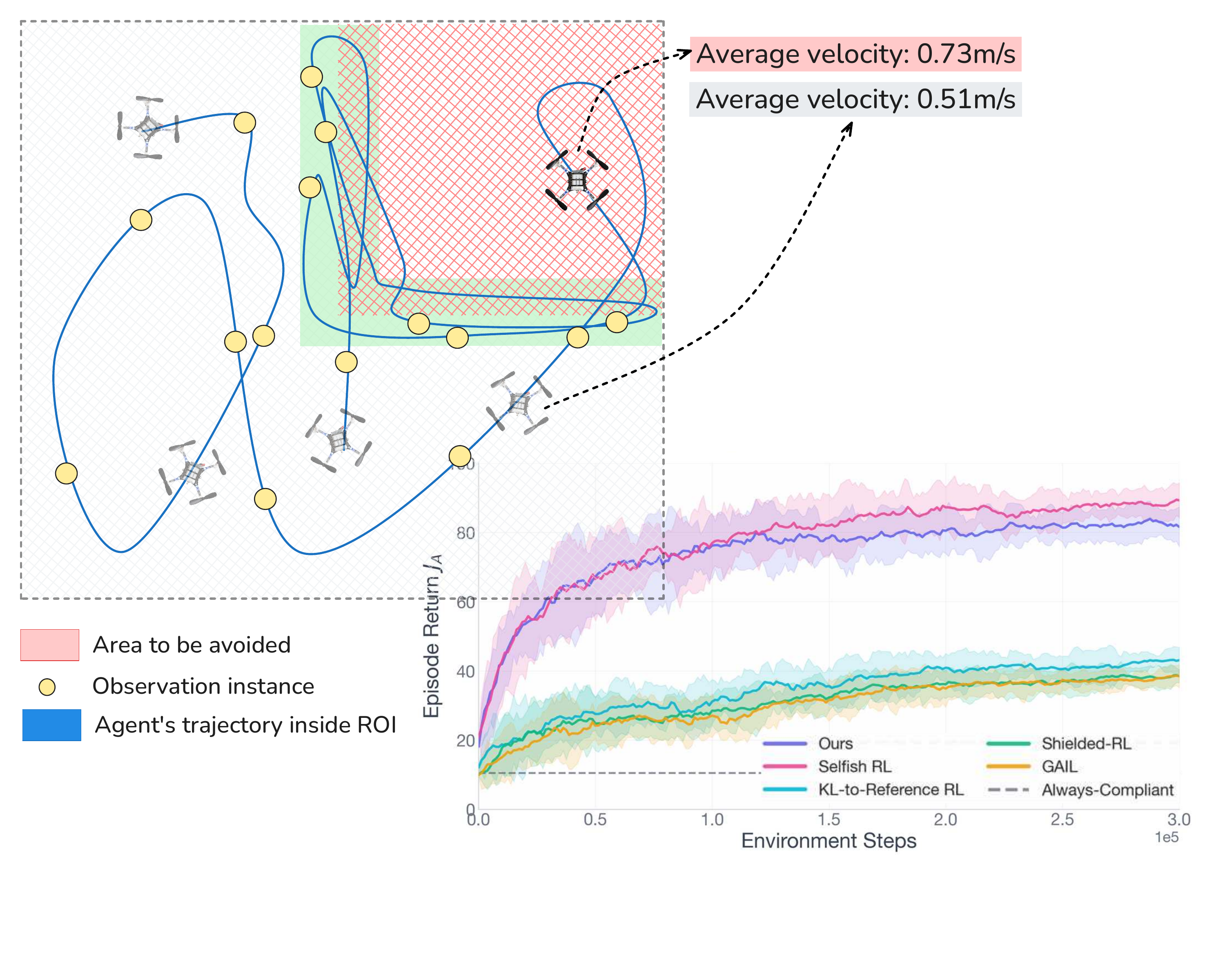}
    \caption{Representative aerial scenario: the supervisor expects the agent to avoid the top right quadrant (red-bounded region).}
    \label{fig:kf4}
\end{figure}

\noindent \tbm{Key Finding \#2: Our method scales seamlessly while remaining low detectability.}
\noindent Moving from 2D marine setting to 3D aerial setting required no bespoke architecture changes: in the aerial setting, our policy achieves $\mathrm{SR}=97/100$ and $J_A=81.9$ (of 100), holds KL at observation at $0.52$ and LLR at $0.54$ with Top-$k=0.83$, and extends TtF to $317$ steps--within $32$ steps of the Always-Compliant ceiling ($349$)--showing that our approach transfers cleanly to higher-dimensional dynamics (refer Table \ref{tab:rl_comparison} for quantitative results and Fig. \ref{fig:kf3} for qualitative results). The slight conservatism (small drop in $J_A$ and SR relative to 2D) reflects tighter control in 3D under more complex dynamics and sparser, occluded observations, where the learned hazard beliefs enforce larger safety margins. Nevertheless, the approach maintains a clear performance–stealth advantage over non-adaptive regularizers, imitation, and hand-crafted switches.
\noindent \tbm{Key Finding \#3: ToM-conditioned RL exhibits creative maneuvers and accelerates online learning vs baselines.}
Beyond aggregate metrics, the policy uncovers behaviors that remain consistent with the supervisor's observations while exploiting structure in the scene. To illustrate one such example, in the UAV avoid-zone task (Fig. \ref{fig:kf4} top), the agent discovers a corridor-staging strategy: it hugs the narrow seam between the no-go region and the permissible area  ((Fig. \ref{fig:kf4}), marked in green), waiting on the boundary so a short pivot can either bypass or access the region without committing to long detours. 
 A separate emergent pattern we observe is an asymmetric speed profile: brief transits across the avoid zone are executed at higher speed, while motion in the permissible area is deliberately slower—compressing time spent in questionable regions and blending where behavior is expected. 
 Independently of these behaviors, our approach also confers a learning advantage in the online setting (Fig. \ref{fig:kf4}, bottom): the ToM signal provides informative feedback under intermittent observations, so within the same interaction time the policy reaches strong returns earlier and with greater stability than non-adaptive baselines.

\begin{figure}[H]
    \centering
    \includegraphics[width=1\linewidth, trim=20 470 26 28,clip]{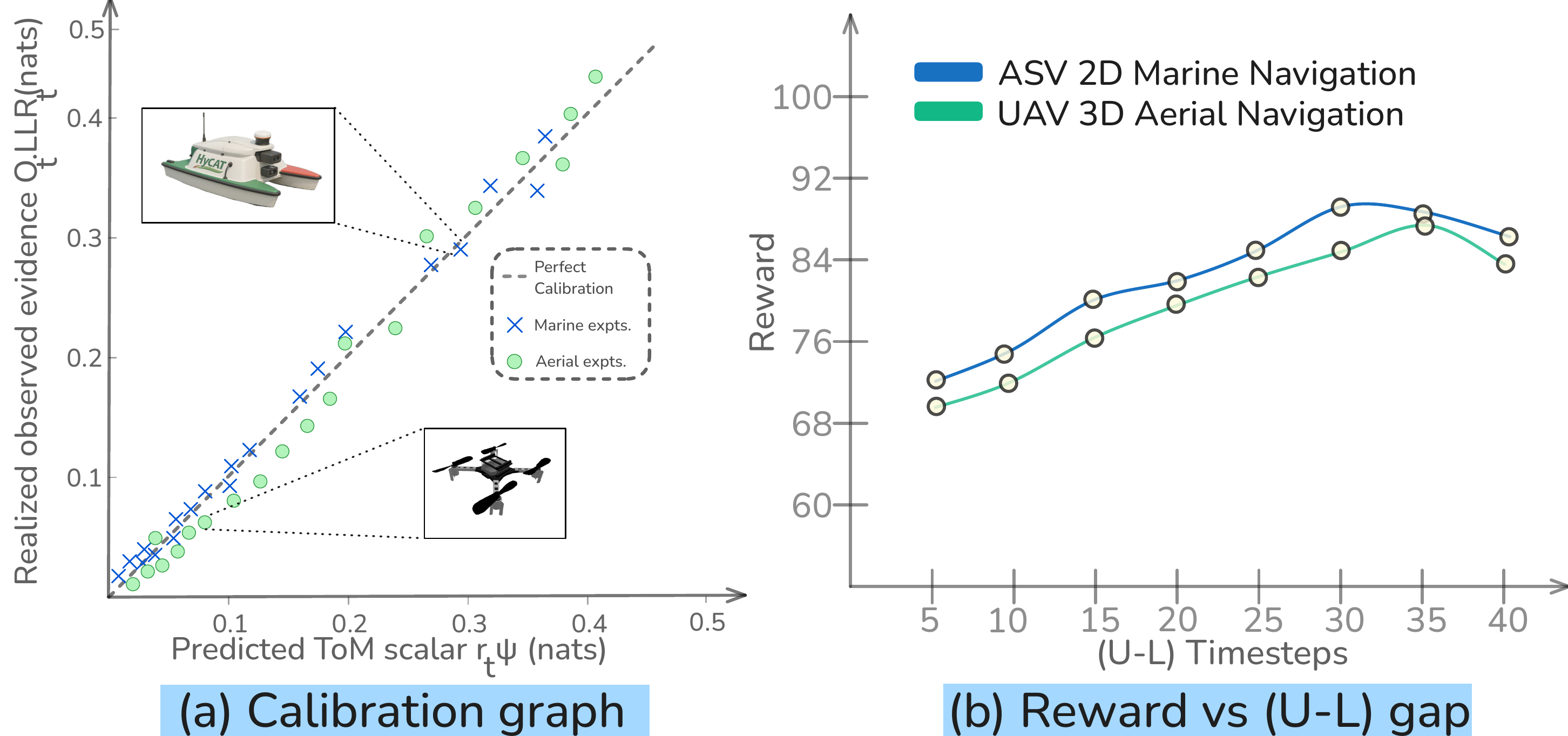}
    \caption{Reliability diagram (20 bins): empirical observed evidential error (y) vs predicted ToM scalar (x). (b) Ablation: reward as a function of the gap length $(U{-}L)$.}
    \label{fig:kf45}
\end{figure}

\noindent \tbm{Key Finding \#4: ToM scalar is numerically reliable and context-discriminative.}
Fig.~\ref{fig:kf45}(a) is a calibration diagram for the ToM scalar. The x-axis is the predicted per-step exposure and the y-axis is the realized observed evidence. The dashed $y{=}x$ line denotes perfect calibration. Points lying on this line mean the exposure we predict is equal, on average, to what the supervisor actually observes. In both domains the cloud of points tightly tracks the identity with small mean bin error and a fitted slope near 1 and intercept near 0, indicating numerical calibration over the range of exposures. Majority of the marine (2D) experiments adhere to the line across low–high $\psi_t$, while aerial (3D) experiments sit slightly below $y{=}x$ at moderate $\psi_t$, reflecting a mild conservative bias.
This pattern follows directly from how $\psi_t$ is formed: it scales the action-deviation term and the state-occupancy shift by the observation belief. The slight conservatism in the 3D setting is consistent with (i) a modest upward bias in $\hat b_t$ at larger ages of the renewal process and/or (ii) small positive bias in the action or occupancy components in higher-dimensional dynamics--both of which push predictions below the $45^\circ$ line and are preferable to risky under-prediction. 
Fig. \ref{fig:kf1} shows these LLR swings along a full lap for qualitative context.

\noindent \tbm{Key Finding \#5: Wider observation gaps are automatically converted into higher reward without retuning.}
Sweeping the gap $U\!-\!L$ (Fig.~\ref{fig:kf45}) shows that ToM-conditioning reallocates compliance in the right places: longer low-hazard stretches are exploited for task progress while short, high-hazard windows trigger tighter, reference-consistent actions. As the window widens, the policy lengthens these exploitation phases and compresses compliance bursts, yielding steadily higher reward and longer undetected operation, while keeping observed evidence bounded. The behavior transfers intact to 3D UAV—slightly more conservative due to harder dynamics and wider safety margins, yet following the same trend—demonstrating that a single, state/time-aware signal is enough to adapt to monitoring cadence. Non-adaptive penalties, hard switches, and fixed blends cannot reallocate effort with the gap and therefore leave performance on the table; our online method turns sparser oversight into sustained gains with no extra tuning.


\section{Conclusions} \label{sec:Conclusions}
We studied deception under intermittent observation as a constrained control problem: maximize private return while keeping the observed behavior statistically close to a supervisor's reference. Our key contribution is a local, calibrated Theory-of-Mind signal $\psi_t$ that multiplies the belief of being observed by a per-state evidence term. Conditioning standard actor–critic updates on $\psi_t$ yields a lightweight online SAC-style algorithm with a state-dependent temperature and a dual ascent that enforces a discounted detectability budget. Across marine (2D) and aerial (3D) platforms, this ToM-conditioned RL preserves near-optimal task success while operating at near–always-compliant stealth, extends time-to-failure, and adapts automatically to wider observation gaps; the ToM scalar is numerically well-calibrated. Taken together, the results show that a simple, per-step evidence signal is sufficient to steer online learning to generate deceptive policies under intermittent observations.

\section*{Acknowledgements}
This work is supported by the Office of Naval Research grant N00014-23-1-2789 and N00014-23-1-2651.

\balance

\bibliographystyle{IEEEtran}
\bibliography{references}

\end{document}